\documentclass[letterpaper, 10 pt, conference]{ieeeconf}  

\IEEEoverridecommandlockouts                              

\usepackage{bm}         
\usepackage{url}  
\usepackage{tabularx}
\usepackage{array}
\usepackage{multirow}
\usepackage{amsmath}
\usepackage{siunitx}
\usepackage{makecell}
\usepackage{amssymb}
\usepackage{xcolor}
\usepackage{graphicx}

\usepackage[normalem]{ulem}

\usepackage[colorlinks=false, linkcolor=blue, citecolor=blue, urlcolor=blue, pdfborder={0 0 0}]{hyperref}       
\hypersetup{pdfborder={0 0 0}}  

\usepackage{booktabs}

\usepackage[normalem]{ulem}

\usepackage{makecell}

\title{\LARGE 
M-CROSS: A Multisensor Simulation Dataset for SLAM \\ across Water--Land Transitions
}

\author{Yan Dong$^{1,2}$, Tianyi Wu$^{2}$, Junyu Yang$^{2}$, Shaoqiang Qiu$^{2}$, Zefeng Yan$^{2,3}$, Tao Zhang$^{1}$ and Bin Han$^{2}$
\thanks{$^{1}$ School of Automation, Tsinghua University, Beijing, China}%
\thanks{$^{2}$ State Key Laboratory of Intelligent Manufacturing Equipment and Technology, School of Mechanical Science and Engineering, Huazhong University of Science and Technology, Wuhan, China}%
\thanks{$^{3}$ School of Automation and Software Engineering, Shanxi University, Shanxi, China}
}

\begin{document}

\maketitle
\thispagestyle{empty}
\pagestyle{empty}

\begin{abstract}
Water--land cross-domain robots require continuous localization across land, transition and underwater phases, where sensor characteristics and availability change abruptly. Existing datasets and conventional metrics are not designed to capture transition-induced degradation and failures. We present \textbf{M-CROSS}, to the best of our knowledge, the first multisensor simulation dataset and benchmark for SLAM across continuous water--land transitions. It provides synchronized LiDAR, sonar, stereo camera, IMU, DVL, and depth measurements; 6DoF pose ground truth; sensor-wise medium-state annotations; and medium-dependent sensor degradation models. We establish a transition-aware protocol comprising the Trajectory Completion Rate (TCR), Piecewise Absolute Pose Error (PW-APE), Transition End Error (TEE), and Transition Survival Period (TSP), which measure trajectory completion, single-domain accuracy, transition-induced error, and localization survival under prolonged partial immersion, respectively. Experiments with 6 representative SLAM systems show that existing methods fail to maintain both localization accuracy and trajectory continuity across the full set of water--land transitions, exhibiting transition-induced drift, trajectory discontinuities, and tracking failures. M-CROSS is publicly available\footnote{\url{https://anonymous.4open.science/r/MCross-Dataset}}.
\end{abstract}


\section{Introduction}

Amphibious robots have demonstrated the ability to operate across terrestrial, water-surface, and underwater environments. Representative platforms include AQUA~\cite{dudek2007aqua}, AmphiHex-I~\cite{zhang2016amphihex}, and the tracked amphibious robot presented in~\cite{yan2023tracked}. These platforms employ flipper-based locomotion, transformable flipper-leg propulsion, and tracked locomotion, respectively. Visual and multimodal perception have also supported underwater tasks such as diver following, robot tracking, inspection, and autonomous exploration \cite{girdhar2015curiosity,shkurti2017convoying,rahman2019svin2}. However, most existing studies focus on locomotion, control, or perception within individual media. Continuous localization across the air--water interface remains insufficiently investigated, although sensor characteristics and availability change substantially during this process.

A major limitation is the lack of datasets designed for continuous cross-domain localization. KITTI \cite{geiger2012kitti}, EuRoC \cite{burri2016euroc}, and TUM RGB-D provide standardized measurements and ground truth for terrestrial, aerial, or indoor environments \cite{sturm2012tum}, while AQUALOC \cite{ferrera2019aqualoc}, FLSea \cite{randall2026flsea}, and AquaticVision \cite{peng2025aquaticvision} focus on underwater visual or visual--inertial perception. These datasets are largely confined to a single physical medium. Real-world cross-domain data are difficult to collect under repeatable conditions, and continuous 6DoF ground truth spanning terrestrial, partially immersed, and underwater phases is particularly challenging to obtain. Simulation offers a controllable alternative: TartanAir supports photorealistic visual SLAM evaluation \cite{wang2020tartanair}, while Stonefish \cite{cieslak2019stonefish}, DAVE \cite{zhang2022dave}, and OceanSim \cite{song2025oceansim} provide underwater optical, acoustic, hydrodynamic, and navigation-sensor simulation. Nevertheless, existing resources do not jointly target continuous land--interface--underwater trajectories with complementary terrestrial and underwater sensors, robot- and sensor-level medium annotations, medium-dependent sensor models, and transition-aware evaluation metrics.

\begin{figure}[t]
    \centering
    \includegraphics[width=0.99\columnwidth]{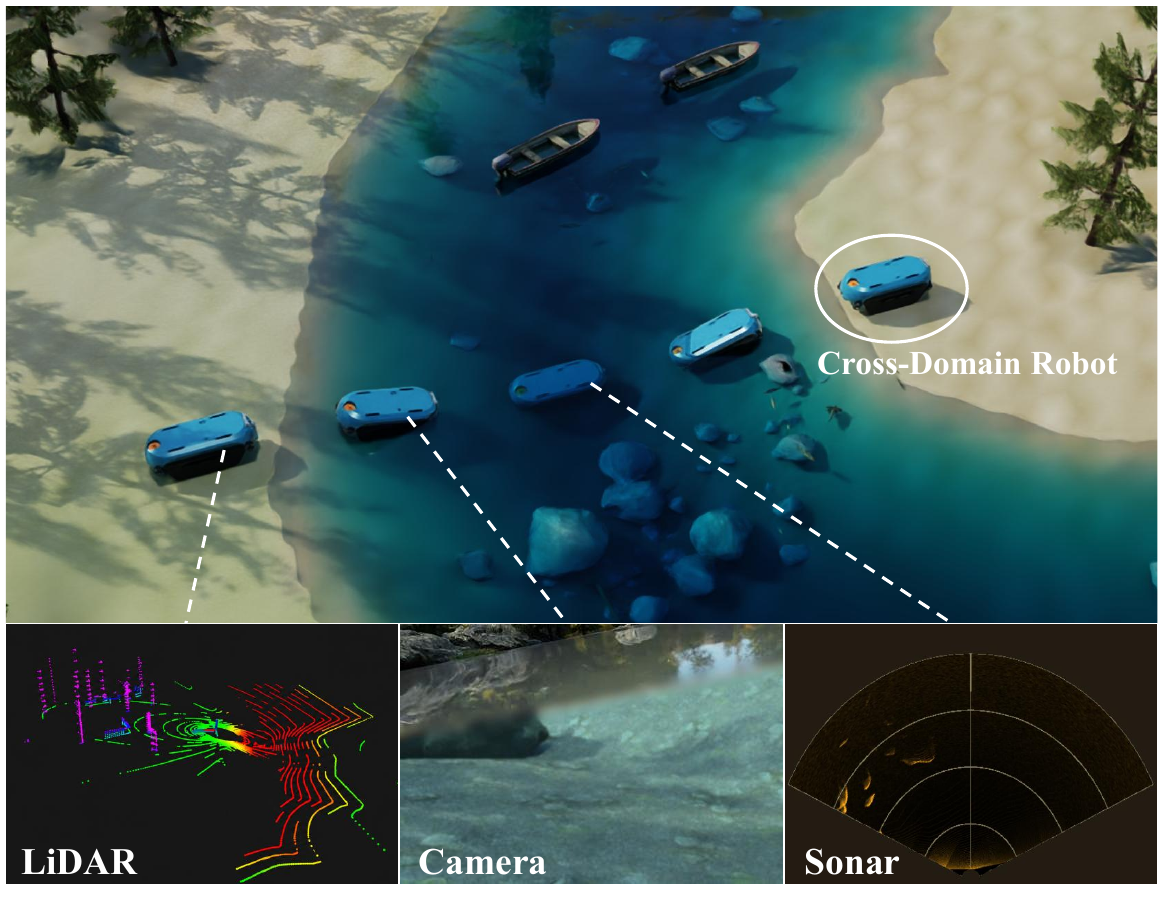}
    \caption{Overview of the M-CROSS simulation scene. The cross-domain robot traverses from one riverbank to the other through continuous land--water transitions. Representative LiDAR, camera, and sonar measurements are shown at the bottom.}
    \label{fig:advertising}
\end{figure}

Evaluation is also challenging. Absolute Pose Error (APE) characterizes global trajectory accuracy \cite{sturm2012tum,zhang2018evaluation}, while existing benchmarks additionally consider cumulative error, incomplete trajectories, and relocalization \cite{engel2016photometric,zhang2023hiltioxford,shi2020openloris}. However, these measures are not explicitly associated with medium-transition events. Errors concentrated within short transition intervals may be diluted by longer single-domain trajectories, and conventional metrics cannot directly characterize how long localization remains reliable during prolonged partial immersion. A cross-domain benchmark must therefore distinguish continuous completion, single-domain accuracy, transition-induced error, and robustness under sustained interface disturbances.

\begin{figure*}[t]
    \centering
    \includegraphics[width=0.99\linewidth]{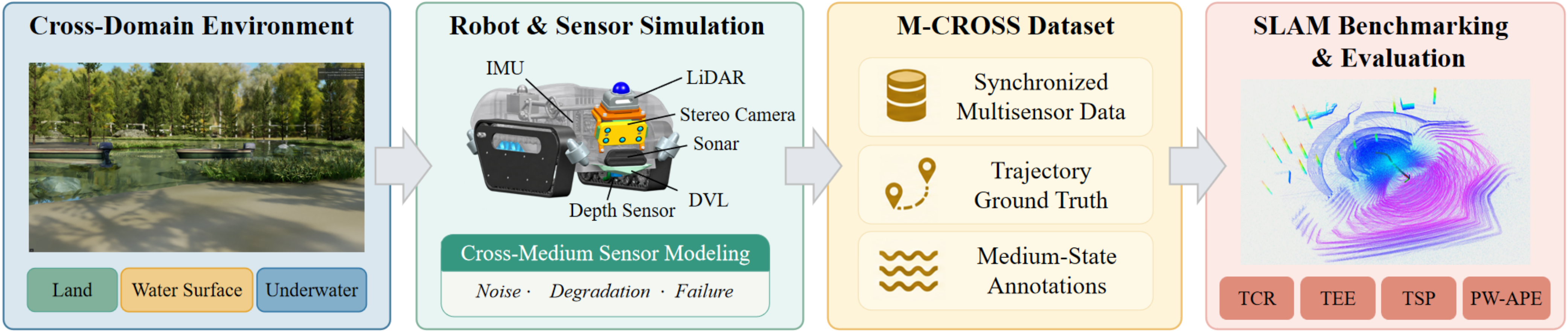}
    \caption{Overview of the dataset construction and benchmarking workflow.}
    \label{fig:Overview}
\end{figure*}

To address these gaps, we present \textbf{M-CROSS}, a multisensor simulation dataset and benchmark for cross-domain water--land SLAM, as illustrated in Fig.~\ref{fig:advertising}. The main contributions are:

\begin{itemize}
    \item We introduce \textbf{M-CROSS}, providing synchronized multimodal measurements, continuous six-degree-of-freedom ground truth, medium-state annotations, and medium-dependent sensor degradation models.

    \item We establish an evaluation framework comprising TCR, PW-APE, TEE, and TSP to separately assess trajectory completion, single-domain accuracy, transition-induced error, and prolonged-interface robustness.

    \item We benchmark representative visual, LiDAR-based, sonar-based, and multisensor SLAM methods, revealing transition-induced drift, trajectory discontinuities, and tracking failures under changing sensing modalities.
\end{itemize}

\section{Simulation Method and M-CROSS Dataset}
As illustrated in Fig.~\ref{fig:Overview}, the overall workflow comprises cross-domain environment construction, robot and multisensor simulation, cross-medium sensor modeling and dataset generation, followed by SLAM benchmarking using transition-aware metrics.

\subsection{Cross-Domain Environment Setup}
We use NVIDIA Isaac Sim to construct the continuous land--interface--underwater environment used in M-CROSS. The scene layout was designed with reference to satellite imagery of real-world streams, preserving representative spatial characteristics such as channel curvature, variations in river width, and irregular shorelines. The environment comprises contiguous terrestrial areas, sloped riverbanks, a water surface, and an underwater riverbed. The terrestrial region contains environmental elements such as trees and rocks, whereas the aquatic region includes both static and dynamic objects, including underwater rocks, a shipwreck, surface vessels, and swimming fish. In addition, the water environment incorporates surface reflections and refraction effects at the air--water interface during rendering, thereby improving the environmental realism of cross-medium sensor data.

\subsection{Cross-Domain Robot and Multisensor Configuration}
The pressure-based depth sensor provides noisy measurements of depth below the water surface when submerged and outputs \texttt{NaN} in air. The main parameters of these sensors are configured with reference to typical specifications of real-world devices, as summarized in Table~\ref{tab:sensor_config}. Among them, the stereo RGB camera, LiDAR, and IMU are implemented using the built-in sensor models in Isaac Sim, while the forward-looking imaging sonar is simulated using the OceanSim plugin~\cite{song2025oceansim,potokar2022holocean}. The DVL is modeled using a four-beam ray-casting-based bottom-tracking approximation, with PhysX used to determine valid riverbed returns and generate velocity measurements from the robot motion relative to the riverbed. Sensor-specific noise models are applied to all modalities.

\subsection{Cross-Medium Sensor Modeling}

\begin{table}[t]
\centering
\caption{Key sensor parameters of M-CROSS}
\label{tab:sensor_config}
\small
\setlength{\tabcolsep}{4pt}
\renewcommand{\arraystretch}{1.08}

\begin{tabularx}{\columnwidth}{
    l
    >{\centering\arraybackslash}p{11mm}
    X
}
\toprule
\textbf{Sensor} & \textbf{Rate} & \textbf{Key Configuration} \\
\midrule

Stereo RGB
& 20 Hz
& $1280\times720$, 0.107 m baseline \\

LiDAR
& 10 Hz
& 32 channels, $360^\circ\times30^\circ$ FoV \\

Sonar
& 5 Hz
& $130^\circ\times20^\circ$ FoV, 0.5--10 m range \\

Depth Sensor
& 5 Hz
& 0--16 bar, 0.01 m depth noise std. \\

DVL
& 5 Hz
& 0.05--125 m bottom-track range, 0.005 m/s velocity noise std. \\

IMU
& 200 Hz
& $60~\mu g/\sqrt{\mathrm{Hz}}$ acc., 
$0.007^\circ/\mathrm{s}/\sqrt{\mathrm{Hz}}$ gyro. noise density \\

\bottomrule
\end{tabularx}
\end{table}

\begin{table*}[t]
\centering
\caption{Overview of the M-CROSS dataset sequences and their
cross-domain traversal patterns.}
\label{tab:dataset_sequences}
\small
\setlength{\tabcolsep}{3.5pt}
\renewcommand{\arraystretch}{1.18}

\begin{tabularx}{\textwidth}{
    >{\centering\arraybackslash}p{1.1cm}
    >{\centering\arraybackslash}p{3.5cm}
    >{\raggedright\arraybackslash}X
    >{\centering\arraybackslash}p{1.2cm}
    >{\centering\arraybackslash}p{1.2cm}
}
\toprule
\textbf{Seq.} &
\textbf{Domain Pattern} &
\textbf{Traversal Description} &
\textbf{Dur. (s)} &
\textbf{Len. (m)} \\
\midrule

\texttt{seq01} &
$\mathrm{L}\xrightarrow{\mathrm{T}}\mathrm{U}$ &
Nominal land-to-underwater entry over flat terrain, with rocks and a
submerged wreck &
39.32 & 45.59 \\

\texttt{seq02} &
$\mathrm{U}\xrightarrow{\mathrm{T}}\mathrm{L}$ &
Nominal underwater-to-land exit over flat terrain, with rocks and a
submerged wreck &
39.89 & 45.05 \\

\texttt{seq03} &
$\mathrm{L}\xrightarrow{\mathrm{T}}\mathrm{U}$ &
Challenging land-to-underwater entry over rough terrain, with rocks,
vegetation, a floating boat, and swimming fish &
61.22 & 69.28 \\

\texttt{seq04} &
$\mathrm{U}\xrightarrow{\mathrm{T}}\mathrm{L}$ &
Challenging underwater-to-land exit over rough terrain, with rocks,
vegetation, a floating boat, and swimming fish &
86.54 & 64.40 \\

\texttt{seq05} &
$\mathrm{L}\xrightarrow{\mathrm{T}}\mathrm{U}
\xrightarrow{\mathrm{T}}\mathrm{L}$ &
Land--underwater--land loop over rough terrain, with trees, rocks,
vegetation, a floating boat, and swimming fish &
103.88 & 117.69 \\

\texttt{seq06} &
$\mathrm{U}\xrightarrow{\mathrm{T}}\mathrm{L}
\xrightarrow{\mathrm{T}}\mathrm{U}$ &
Underwater--land--underwater loop over rough terrain, with trees,
rocks, vegetation, a floating boat, and swimming fish &
87.38 & 101.90 \\

\texttt{seq07} &
$\mathrm{L}\overset{\mathrm{T}}{\rightleftarrows}\mathrm{U}$
(\(5\) crossings) &
Five alternating land-to-underwater and underwater-to-land crossings over rough terrain,
with trees, rocks, vegetation, a submerged wreck, and swimming fish &
109.82 & 124.42 \\

\texttt{seq08} &
$\mathrm{L}\xrightarrow{\mathrm{T}^{\ast}}\cdots$ &
Repeated partial immersion near the air--water interface, starting
from land without completing a full crossing &
110.18 & 130.15 \\

\bottomrule
\end{tabularx}

\vspace{1mm}
\begin{minipage}{0.98\textwidth}
\scriptsize
\textit{Note:}
$\mathrm{L}$ and $\mathrm{U}$ denote the Land and Underwater phases,
respectively, while $\mathrm{T}$ above each arrow denotes the
intervening Transition phase. The bidirectional arrow for
\texttt{seq07} denotes five alternating crossings, starting on land
and ending underwater. In \texttt{seq08}, $\mathrm{T}^{\ast}$ denotes
a prolonged-interface phase without a completed crossing.
\end{minipage}
\end{table*}


\begin{figure*}
    \centering
    \includegraphics[width=0.9\linewidth]{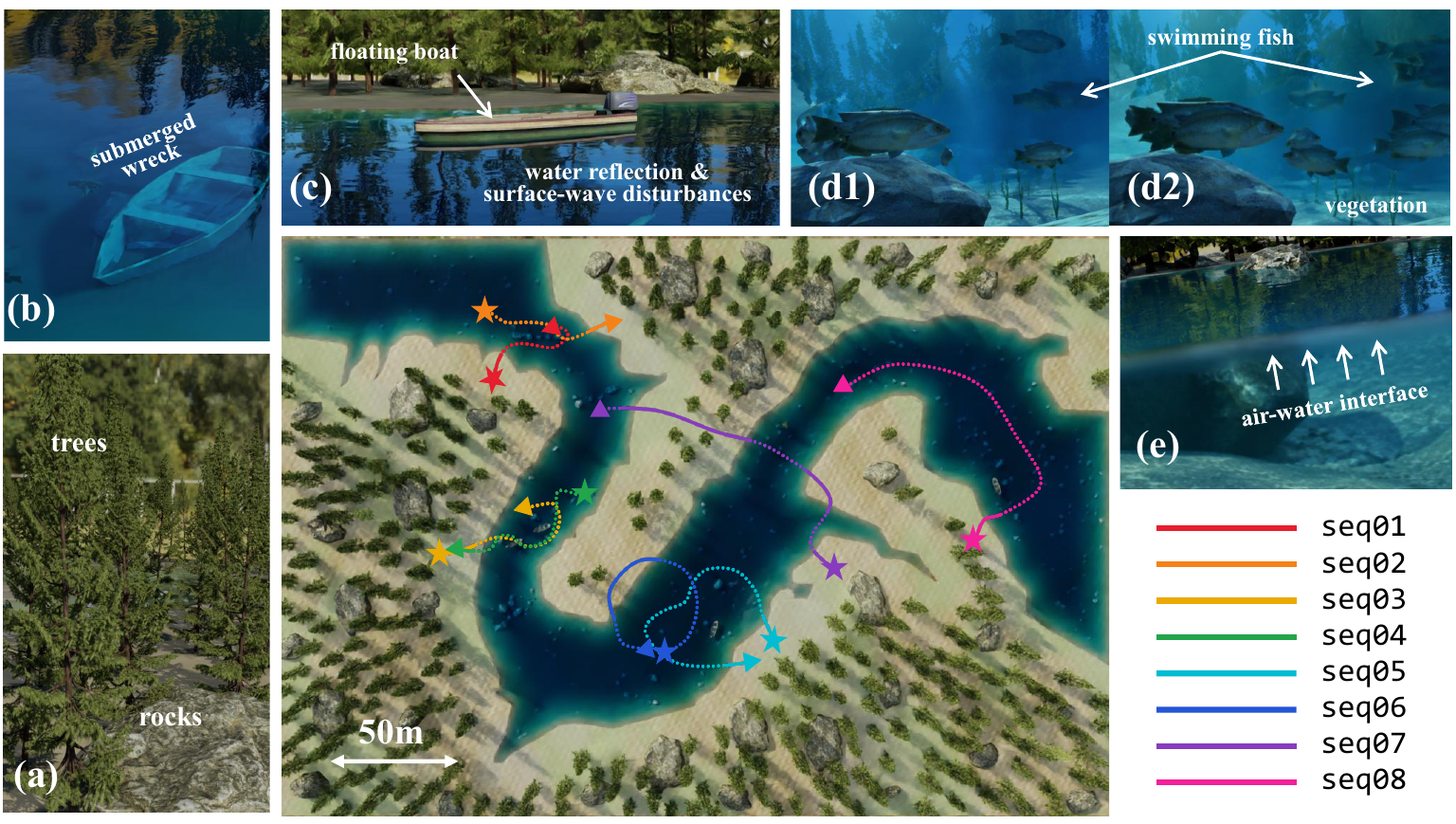}
    \caption{Overview of the M-CROSS environment and sequence trajectories. The central panel shows the overall simulation environment and the trajectories of all eight sequences, while panels (a)--(e) present representative views of different regions and environmental elements. Star and triangle markers indicate the starting and ending points of the trajectories, respectively. For each sequence, trajectory segments above the water surface are shown as solid lines, whereas underwater segments are shown as dashed lines.}
    \label{fig:dataset_overview}
\end{figure*}

M-CROSS models the medium-dependent degradation, failure, and recovery of each sensor according to its sensing state relative to the water surface, while maintaining continuous message streams and timestamps across the in-air, transition, and underwater stages. These models are intended to reproduce the dominant cross-medium failure patterns rather than fully simulate the underlying optical and acoustic propagation.

For the stereo RGB camera, turbidity-dependent contrast attenuation and a blue--green color shift are applied to approximate the visibility degradation caused by underwater absorption and scattering. Near the air--water interface, refraction and water-surface disturbances may introduce locally non-rigid image distortions. Based on the existing refraction rendering, M-CROSS therefore applies spatially varying blur around the projected waterline.

The LiDAR operates nominally in air. When laser beams intersect the water surface, water-surface returns are probabilistically dropped according to the incidence angle and surface conditions, while additional range noise is added to the retained measurements~\cite{yan2019slier}. Once fully submerged, near-infrared laser measurements are strongly affected by absorption, scattering, and backscattering~\cite{castillon2019state}. M-CROSS therefore suppresses valid returns from underwater targets and generates short-range random measurements along the original scanning directions, approximating the near-field invalid returns observed in preliminary experiments.

The forward-looking imaging sonar produces valid range--bearing intensity images underwater but only background noise in air. The bottom-tracking DVL provides valid measurements when submerged and a sufficient number of beams return valid bottom echoes; otherwise, its output is encoded as \texttt{NaN}. The pressure-based depth sensor provides noisy measurements of depth below the water surface when submerged and outputs \texttt{NaN} in air. The IMU operates continuously without additional medium-dependent degradation. When a sensor returns to a medium in which it operates nominally, its standard measurement model is resumed.

\subsection{Dataset Sequences}

M-CROSS contains eight continuous water--land transition sequences, as
summarized in Table~\ref{tab:dataset_sequences}. \texttt{seq01}--\texttt{seq04}
contain single crossings in opposite directions under nominal or
challenging conditions. \texttt{seq05} and \texttt{seq06} form
complementary two-crossing loops, while \texttt{seq07} contains five
alternating complete crossings. \texttt{seq08} instead maintains
prolonged partial immersion without completing a crossing.

Figure~\ref{fig:dataset_overview} shows the simulation environment, all
eight trajectories, and representative scene regions.

\subsection{Ground Truth and Data Organization}

M-CROSS provides time-stamped six-degree-of-freedom ground-truth trajectories for the robot body and all onboard sensors. Sensor poses are derived from the robot-body trajectory using fixed extrinsic calibrations. The dataset also provides sensor-wise medium-state annotations indicating whether each sensor is in air or water, as determined from its position and the water-region information encoded in the simulation map. Because the sensors are mounted at different positions, their medium states may change at different times during an air--water transition.

All measurements retain their native sampling rates and are timestamped using a shared simulation clock, supporting synchronized playback and trajectory evaluation.

\section{Cross-Domain Localization Evaluation Metrics}
\label{sec:evaluation_metrics}

We use Trajectory Completion Rate (TCR) to evaluate sequence completion,
Piecewise Absolute Pose Error (PW-APE) for single-domain accuracy,
Transition End Error (TEE) for transition-induced error, and Transition
Survival Period (TSP) for robustness under prolonged interface
conditions.

\subsection{Definition of Water--Land Transition Phases}
\label{subsec:transition_definition}

A cross-domain event begins when the robot leaves one single-domain
phase and ends when it fully enters the other medium. The robot is in
the Land phase when entirely above water, the Underwater phase when
fully submerged, and the Transition phase when intersecting the
air--water interface. Land and Underwater are termed single-domain
phases. These robot-level phases differ from sensor-wise
\emph{air}/\emph{water} states because individual sensors cross the
interface at different times.

\begin{figure}[t]
    \centering
    \includegraphics[width=0.99\linewidth]{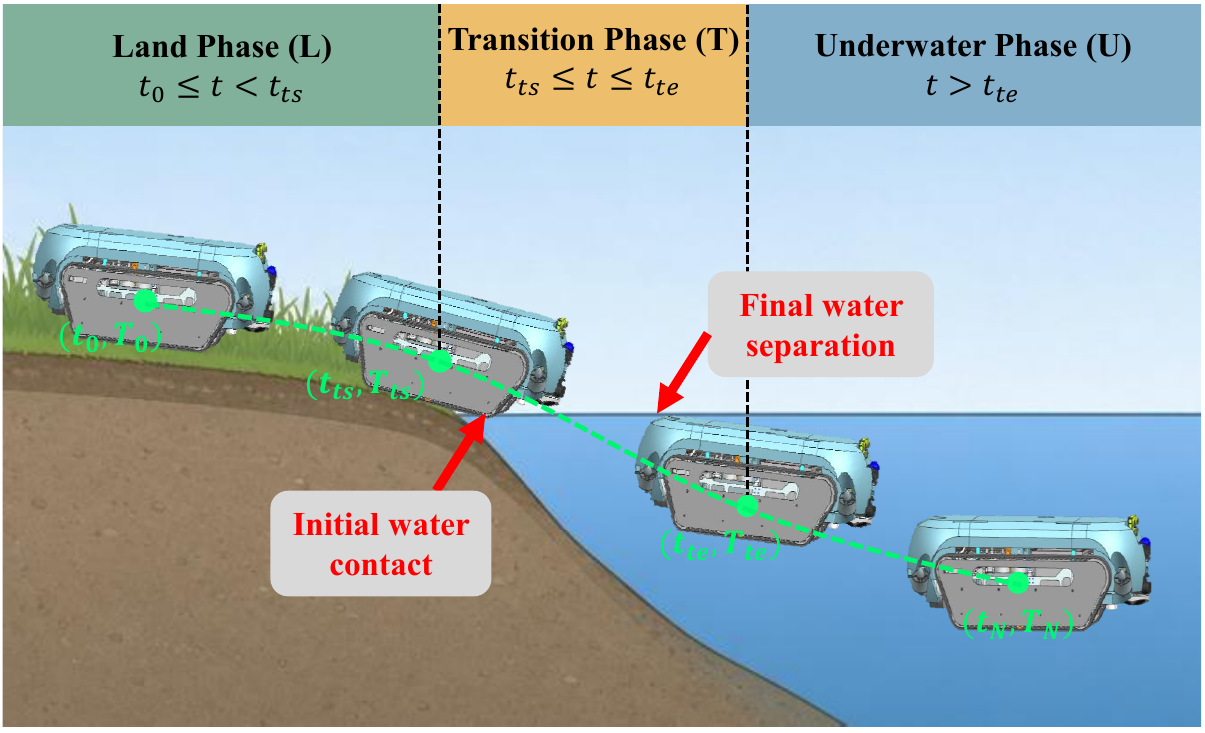}
    \caption{Definition of cross-domain phases. A completed Transition
    spans the first body--surface intersection ($t_{ts}$) to full entry
    into the target medium ($t_{te}$); an incomplete crossing remains
    in the Transition phase until the annotated endpoint $t_c$.}
    \label{fig:transition_definition}
\end{figure}

As shown in Fig.~\ref{fig:transition_definition}, the $k$-th completed
transition spans $[t_{ts}^{(k)},t_{te}^{(k)}]$, with boundaries
determined by the robot geometry and simulated water surface rather than
individual sensor states. Successive Land, Transition, and Underwater
phases are denoted by $\mathrm{L}_i$, $\mathrm{T}_i$, and
$\mathrm{U}_i$. An incomplete crossing remains a Transition phase and
has no corresponding $t_{te}^{(k)}$.

All methods are initialized using the same preceding data and operate
continuously over $[t_0,t_N]$. Initialization data are excluded, and
estimated poses are transformed into the robot body frame and matched
with ground truth at common timestamps. Localization fails upon reported
tracking loss, abnormal termination, a pose gap exceeding
$\tau_{\mathrm{gap}}=1~\mathrm{s}$, or translational error exceeding
$d_{\mathrm{fail}}=5~\mathrm{m}$ after a one-time $\mathrm{SE}(3)$
alignment at $t_0$. Outputs after the first failure are discarded, with
no relocalization or reinitialization accepted.

\subsection{Evaluation Protocol and Trajectory Completion Rate}
\label{subsec:evaluation_protocol}

Let $t_f$ be the last timestamp of the continuously valid trajectory before the first failure. We set $t_f=t_N$ if the method remains valid throughout the sequence and $t_f=t_0$ if initialization is not completed. The Trajectory Completion Rate is defined as
\begin{equation}
    \mathrm{TCR}=\frac{t_f-t_0}{t_N-t_0}\times 100\%,
    \label{eq:tcr}
\end{equation}
which measures the continuously completed proportion of the sequence.

Accuracy metrics are evaluated on individual annotated intervals rather than being conditioned on full-sequence completion. An interval containing the first failure is excluded, whereas previously completed intervals remain valid. The number of valid intervals is reported with each metric, and \textit{N/A} is used when none is available. For aggregate reporting, each method is evaluated over its own valid intervals. Because these interval sets differ across methods, PW-APE and TEE must be interpreted jointly with their reported valid-interval coverage rather than compared in isolation.

\subsection{Piecewise Absolute Pose Error}
\label{subsec:piecewise_ape}

PW-APE measures translational and rotational accuracy within single-domain phases. After excluding all Transition phases, a sequence is divided into $J$ continuous Land or Underwater segments. Each segment is independently aligned in $\mathrm{SE}(3)$ without scale correction. Following the standard APE definition~\cite{grupp2017evo}, the translational and rotational RMSEs of the $j$-th segment are denoted by $\mathrm{APE}_{P,j}$ and $\mathrm{APE}_{R,j}$.

Let $\mathcal{J}_{\mathrm{val}}$ contain the single-domain segments that are continuously and completely traversed, and let $N_j$ be the number of evaluation poses in segment $j$. To account for different segment durations, PW-APE is computed as the pooled RMSE
\begin{equation}
\scalebox{1}{$
\mathrm{PW\text{-}APE}_{\xi}
=
\sqrt{\frac{\sum_{j\in\mathcal{J}_{\mathrm{val}}}N_j \cdot \mathrm{APE}_{\xi,j}^{2}}{\sum_{j\in\mathcal{J}_{\mathrm{val}}}N_j}},
\qquad
\xi\in\{P,R\}.
$}
\label{eq:pw_ape}
\end{equation}
Here, $\mathrm{PW\text{-}APE}_{P}$ and $\mathrm{PW\text{-}APE}_{R}$ are reported in meters and degrees, respectively. Independent segment alignment isolates single-domain estimation accuracy; it is used only for offline evaluation and does not imply reinitialization at phase boundaries. The valid-segment count $J_{\mathrm{val}}$, together with the total number of single-domain segments $J$, is reported with PW-APE.

\subsection{Transition End Error}
\label{subsec:tee}

TEE measures the net relative pose error accumulated over a complete cross-domain event. For the $k$-th transition, using $t_{ts}^{(k)}$ as the fixed reference time, the transition-relative pose error is
\begin{equation}
    \boldsymbol{E}_{\mathrm{tr}}^{(k)}(t)=\left[\boldsymbol{T}(t_{ts}^{(k)})^{-1}\boldsymbol{T}(t)\right]^{-1}\left[\hat{\boldsymbol{T}}(t_{ts}^{(k)})^{-1}\hat{\boldsymbol{T}}(t)\right],
    \label{eq:transition_error}
\end{equation}
where $\boldsymbol{T}(t)\equiv\boldsymbol{T}_{WB}(t)$ and $\hat{\boldsymbol{T}}(t)$ are the ground-truth and estimated robot-body poses. This formulation removes the rigid-body offset at the transition start without resetting the localization system.

The translational and rotational errors at the transition endpoint are
\begin{equation}
    \begin{aligned}
        \mathrm{TEE}_{P}^{(k)}&=\left\|\operatorname{trans}\!\left(\boldsymbol{E}_{\mathrm{tr}}^{(k)}(t_{te}^{(k)})\right)\right\|_2,\\
        \mathrm{TEE}_{R}^{(k)}&=\operatorname{angle}\!\left(\operatorname{rot}\!\left(\boldsymbol{E}_{\mathrm{tr}}^{(k)}(t_{te}^{(k)})\right)\right).
    \end{aligned}
    \label{eq:tee_event}
\end{equation}
They are reported in meters and degrees, respectively. TEE is averaged over all transition events completed continuously with valid estimates at both endpoints, with all valid transition events weighted equally. The valid-event count $K_{\mathrm{val}}$, together with the total number of TEE-eligible events $K$, is reported with the averaged TEE. Events containing the first tracking failure are excluded.

\subsection{Transition Survival Period}
\label{subsec:tsp}

TSP evaluates how long localization remains within a prescribed position-error threshold during a prolonged Transition phase. Let $t_c$ be the end of the annotated observation interval and let $\boldsymbol{E}_{\mathrm{tr}}(t)$ denote the transition-relative error defined in Eq.~\eqref{eq:transition_error}. For a position-error threshold $\alpha$, the first threshold-crossing time is
\begin{equation}
    t_{\alpha}=\inf\left\{t\in[t_{ts},t_c]\ \middle|\ \left\|\operatorname{trans}\!\left(\boldsymbol{E}_{\mathrm{tr}}(t)\right)\right\|_2>\alpha\right\},
    \label{eq:tsp_threshold}
\end{equation}
where $t_{\alpha}=+\infty$ if the threshold is not exceeded. Combining this crossing time with the first localization failure time $t_f$, TSP is defined as
\begin{equation}
    \mathrm{TSP}_{\alpha}=\min(t_{\alpha},t_f,t_c)-t_{ts}.
    \label{eq:tsp}
\end{equation}

We evaluate $\alpha\in\{0.5,1,2\}~\mathrm{m}$ and use $\mathrm{TSP}_{1}$ as the primary result. If the method remains valid and below the threshold until $t_c$, the result is right-censored and reported as $\mathrm{TSP}_{\alpha}\geq t_c-t_{ts}$. If the method does not continuously reach $t_{ts}$ or has no valid pose at the transition start, TSP is reported as \textit{N/A}. In M-CROSS, TSP is evaluated on \texttt{seq08}.

\begin{table*}[t]
\centering
\caption{Trajectory completion rate of six SLAM methods on all eight M-CROSS sequences.}
\label{tab:tcr_all_sequences}
\begin{tabular}{lcccccc}
\toprule
Seq.
& \mbox{Fast-LIO2~\cite{xu2022fastlio2}}
& \mbox{Bruce-SLAM~\cite{bruceslam}}
& \mbox{ORB-SLAM3~\cite{campos2021orbslam3}}
& \mbox{VINS-Fusion~\cite{qin2018vinsmono}}
& \mbox{LVI-SAM~\cite{shan2021lvisam}}
& \mbox{Fast-LIVO2~\cite{fastlivo2}} \\
\midrule

\texttt{seq01}
& 36.75\% (U$_1$)
& 0.00\% (I)
& 9.48\% (L$_1$)
& \textbf{100.00\%}
& \underline{48.53\% (U$_1$)}
& 28.18\% (T$_1$) \\

\texttt{seq02}
& 12.90\% (U$_1$)
& \textbf{77.39\% (T$_1$)}
& 16.91\% (U$_1$)
& \underline{17.36\% (U$_1$)}
& 10.19\% (U$_1$)
& 0.00\% (I) \\

\texttt{seq03}
& 0.00\% (I)
& 0.00\% (I)
& 20.53\% (T$_1$)
& \textbf{100.00\%}
& \underline{57.17\% (U$_1$)}
& 24.75\% (T$_1$) \\

\texttt{seq04}
& 4.58\% (U$_1$)
& 0.00\% (I)
& 8.73\% (U$_1$)
& \textbf{100.00\%}
& \underline{10.28\% (U$_1$)}
& 0.00\% (I) \\

\texttt{seq05}
& \underline{13.84\% (U$_1$)}
& 0.00\% (I)
& 5.92\% (T$_1$)
& \textbf{100.00\%}
& 13.37\% (U$_1$)
& 9.19\% (T$_1$) \\

\texttt{seq06}
& 5.62\% (U$_1$)
& \underline{29.35\% (U$_1$)}
& 20.25\% (U$_1$)
& \textbf{100.00\%}
& 4.47\% (U$_1$)
& 0.00\% (I) \\

\texttt{seq07}
& 14.19\% (U$_1$)
& 0.00\% (I)
& 7.51\% (T$_1$)
& \textbf{100.00\%}
& \underline{14.49\% (U$_1$)}
& 12.37\% (U$_1$) \\

\texttt{seq08}
& \textbf{40.63\% (T$_1$)}
& 0.00\% (I)
& 10.51\% (T$_1$)
& \underline{32.46\% (T$_1$)}
& 13.71\% (T$_1$)
& 12.58\% (T$_1$) \\

\bottomrule
\end{tabular}

\vspace{1mm}
\begin{minipage}{0.83\textwidth}
\scriptsize
\textit{Note:}
"L$_i$", "T$_i$", and "U$_i$" indicate the first tracking failure in the
$i$-th land, transition, and underwater phase.
"I" denotes initialization failure.
\end{minipage}
\end{table*}

\begin{figure*}[t]
    \centering
    \includegraphics[width=0.92\linewidth]{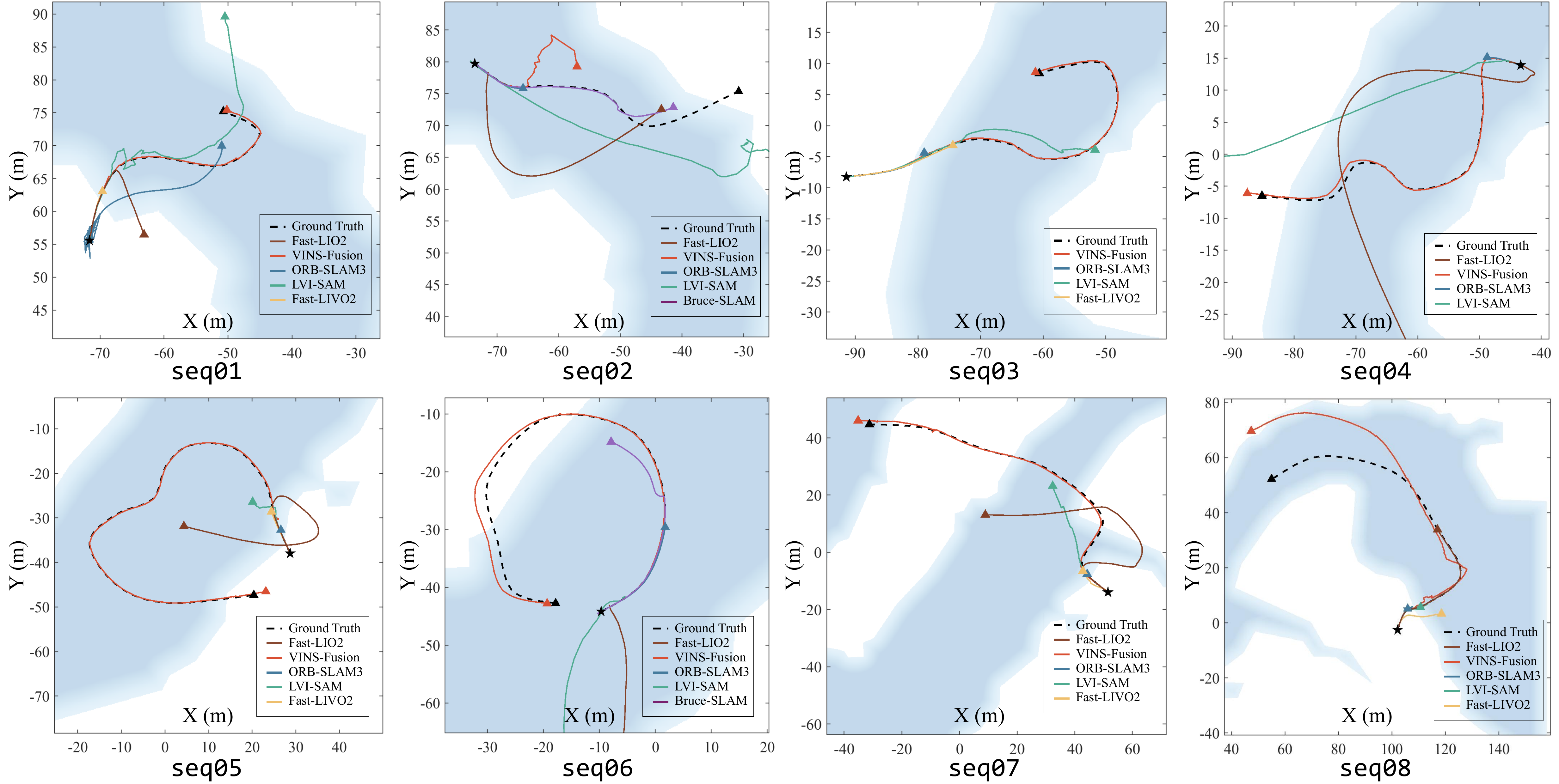}
    \caption{Trajectory comparison of the evaluated SLAM methods on all eight M-CROSS sequences. Each subfigure shows top-down 2-D projections of the estimated and ground-truth trajectories. Star and triangle markers indicate the starting and ending points of the trajectories, respectively.}
    \label{fig:trajectory_comparison}
\end{figure*}

\begin{table*}[t]
\centering
\caption{Dataset-level summary of \textbf{PW-APE} and mean \textbf{TEE} on M-CROSS.}
\label{tab:pw_ape_tee_all_sequences}
\begin{tabular}{@{}lcccc@{}}
\toprule
Method & \multicolumn{2}{c}{Single-domain performance} & \multicolumn{2}{c}{Transition-phase performance} \\
\cmidrule(lr){2-3}\cmidrule(lr){4-5}
& Valid SD-Seg.$^{*}$ $\uparrow$ & $\mathrm{PW\text{-}APE}_{P/R}$ (m/$^{\circ}$) $\downarrow$
& Valid Trans.$^{*}$ $\uparrow$ & $\mathrm{TEE}_{P/R}$ (m/$^{\circ}$) $\downarrow$ \\
\midrule
\mbox{Fast-LIO2~\cite{xu2022fastlio2}}       & 7               & \textbf{0.57} / 1.03     & 1              & \underline{1.03} / 2.07 \\
\mbox{Bruce-SLAM~\cite{bruceslam}}           & 1               & \underline{0.56} / \underline{0.78}  & 0              & \textit{N/A} / \textit{N/A} \\
\mbox{ORB-SLAM3~\cite{campos2021orbslam3}}   & 4               & {0.69} / \textbf{0.54}                         & 1              & 3.10 / 21.21 \\
\mbox{VINS-Fusion~\cite{qin2018vinsmono}}    & \textbf{19}     & 1.03 / 1.17               & \textbf{12}    & \textbf{0.25} / \underline{0.69} \\
\mbox{LVI-SAM~\cite{shan2021lvisam}}         & \underline{9}   & 0.96 / 1.10               & \underline{2}  & 1.80 / \textbf{0.30} \\
\mbox{Fast-LIVO2~\cite{fastlivo2}}           & 2               & 3.18 / 15.90              & 1              & 3.93 / 13.87 \\
\bottomrule
\end{tabular}

\vspace{1mm}
\begin{minipage}{0.78\textwidth}
\scriptsize
\textit{Note:}
Each P/R entry reports the translational and rotational results, respectively.
M-CROSS contains 21 single-domain land/underwater segments and 14 transition phases.
Valid SD-Seg.$^{*}$ denotes the number of single-domain segments with valid PW-APE results.
Valid Trans.$^{*}$ denotes the number of completed transitions with valid endpoint estimates among the 13 TEE-eligible events; the remaining prolonged transition is evaluated using TSP.
PW-APE is computed as the pooled RMSE over the corresponding valid
single-domain segments, whereas TEE is averaged over the valid
transition events.
Bold and underlined values indicate the best and second-best results, respectively.
\textit{N/A} indicates that no valid result is available.
\end{minipage}
\end{table*}

\section{Benchmark Evaluation and Analysis}

\subsection{Experimental Setup}

To evaluate existing SLAM methods under continuous water--land transitions, we test six representative open-source systems: the LiDAR--inertial method Fast-LIO2~\cite{xu2022fastlio2}, the sonar-based method Bruce-SLAM~\cite{bruceslam}, the visual method ORB-SLAM3~\cite{campos2021orbslam3}, the visual--inertial method VINS-Fusion~\cite{qin2018vinsmono}, and the multisensor methods LVI-SAM~\cite{shan2021lvisam} and Fast-LIVO2~\cite{fastlivo2}. Since Bruce-SLAM estimates a planar 3-DoF pose, its vertical position is supplemented using the pressure-derived depth measurements provided by M-CROSS for trajectory evaluation.

Each method uses its required sensor modalities from the synchronized measurements provided by M-CROSS. Except for sensor-parameter and data-interface adaptation, the recommended configurations of the open-source implementations are used without sequence-specific tuning. All sequences are evaluated using TCR, PW-APE, and TEE where applicable, while \texttt{seq08} is additionally used for TSP evaluation.

\subsection{Cross-Domain Completion and Localization Accuracy}
Table~\ref{tab:tcr_all_sequences} reports the TCR of the six methods, while Fig.~\ref{fig:trajectory_comparison} shows their corresponding 2-D trajectories. In addition to completion rate, the table identifies the phase in which the first failure occurs. L, T, and U denote land, transition, and underwater phases, respectively, and the subscript indicates the chronological occurrence of that phase. Initialization failures are assigned a TCR of 0\%.

VINS-Fusion achieves the highest overall completion, producing complete trajectories on \texttt{seq01} and \texttt{seq03}--\texttt{seq07}. It nevertheless fails in U$_1$ on \texttt{seq02} and in T$_1$ on the prolonged-transition sequence \texttt{seq08}. Fast-LIO2 fails on most sequences during the first water-entry or underwater phase, or entering the water before initialization completes, consistent with the modeled degradation of near-infrared LiDAR after water entry. Bruce-SLAM initializes on only two of the three underwater-starting sequences, \texttt{seq02} and \texttt{seq06}; it fails to initialize on the more challenging \texttt{seq04} and on all land-starting sequences.

LVI-SAM generally maintains tracking longer than Fast-LIVO2, although both methods eventually fail. LVI-SAM retains interacting visual--inertial and LiDAR--inertial subsystems, which may provide limited redundancy when one sensing modality degrades. In contrast, Fast-LIVO2 fuses visual and LiDAR observations through a unified state and map, making its estimation more sensitive to abrupt degradation of the LiDAR-supported geometry. These results show that multisensor fusion alone does not ensure continuous cross-domain localization, and that different fusion architectures may exhibit substantially different failure behavior.

Because most methods terminate early on multiple sequences, conventional full-sequence APE would introduce substantial coverage bias. Table~\ref{tab:pw_ape_tee_all_sequences} therefore summarizes the results over annotated segments and transition events rather than complete sequences. M-CROSS contains 21 single-domain segments and 14 transition phases, of which 13 complete transitions are eligible for TEE evaluation; the prolonged transition in \texttt{seq08} is evaluated separately using TSP.

VINS-Fusion provides valid estimates for 19 single-domain segments and 12 transition events, achieving substantially broader coverage than the other methods and the lowest translational TEE. Fast-LIO2 achieves low translational and rotational PW-APE over its seven valid single-domain segments but completes only one transition event. Bruce-SLAM reports slightly lower PW-APE values, although they are obtained from only one valid single-domain segment and should therefore be interpreted with caution. LVI-SAM achieves the lowest rotational TEE, although this value is computed from only two valid events. Cross-domain localization performance must therefore be assessed jointly in terms of estimation accuracy and valid coverage.

\subsection{Robustness under Prolonged Interface Conditions}

\begin{table}[t]
\centering
\caption{\textbf{TEP} under different position-error thresholds on \texttt{seq08}.}
\label{tab:tsp}
\begin{tabular}{lccc}
\toprule
Method
& TSP$_{0.5}$
& TSP$_{1}$
& TSP$_{2}$ \\
\midrule

\mbox{Fast-LIO2~\cite{xu2022fastlio2}}
& 7.59 & 33.90 & 33.90 \\

\mbox{ORB-SLAM3~\cite{campos2021orbslam3}}
& 0.10 & 0.33 & 1.20 \\

\mbox{VINS-Fusion~\cite{qin2018vinsmono}}
& 0.87 & 1.41 & 1.49 \\

\mbox{LVI-SAM~\cite{shan2021lvisam}}
& 3.96 & 4.02 & 4.14 \\

\mbox{Fast-LIVO2~\cite{fastlivo2}}
& 0.87 & 1.74 & 3.21 \\

\bottomrule
\end{tabular}

\vspace{1mm}
\begin{minipage}{0.90\columnwidth}
\scriptsize
\textit{Note:}
All TSP values are in seconds. The subscripts 0.5, 1, and 2 denote
position-error thresholds in meters. Higher values indicate longer
localization survival under prolonged interface conditions.
Bruce-SLAM is omitted because it fails to initialize on \texttt{seq08}.
\end{minipage}
\end{table}

The \texttt{seq08} sequence is designed to evaluate localization robustness under prolonged partial immersion, as may occur during surface maneuvering or shallow-water operation. Table~\ref{tab:tsp} reports the TSP results under different position-error thresholds. None of the evaluated methods completes the entire sequence (\ref{tab:tsp}). Fast-LIO2 achieves the longest survival period. This result is mainly attributable to the high mounting position of the
LiDAR, which remains above the water during the early part of the
transition, rather than an ability to operate after LiDAR immersion.
For the remaining methods, TSP$_2$ does not exceed 4.14~s. Although
cameras can provide images in both air and water, reflection,
refraction, water-surface fluctuation, and mixed-medium views make
visual localization unstable during prolonged interface operation.

\subsection{Transition-Induced Error Analysis}

\begin{figure}[t]
    \centering
    \includegraphics[width=0.99\linewidth]{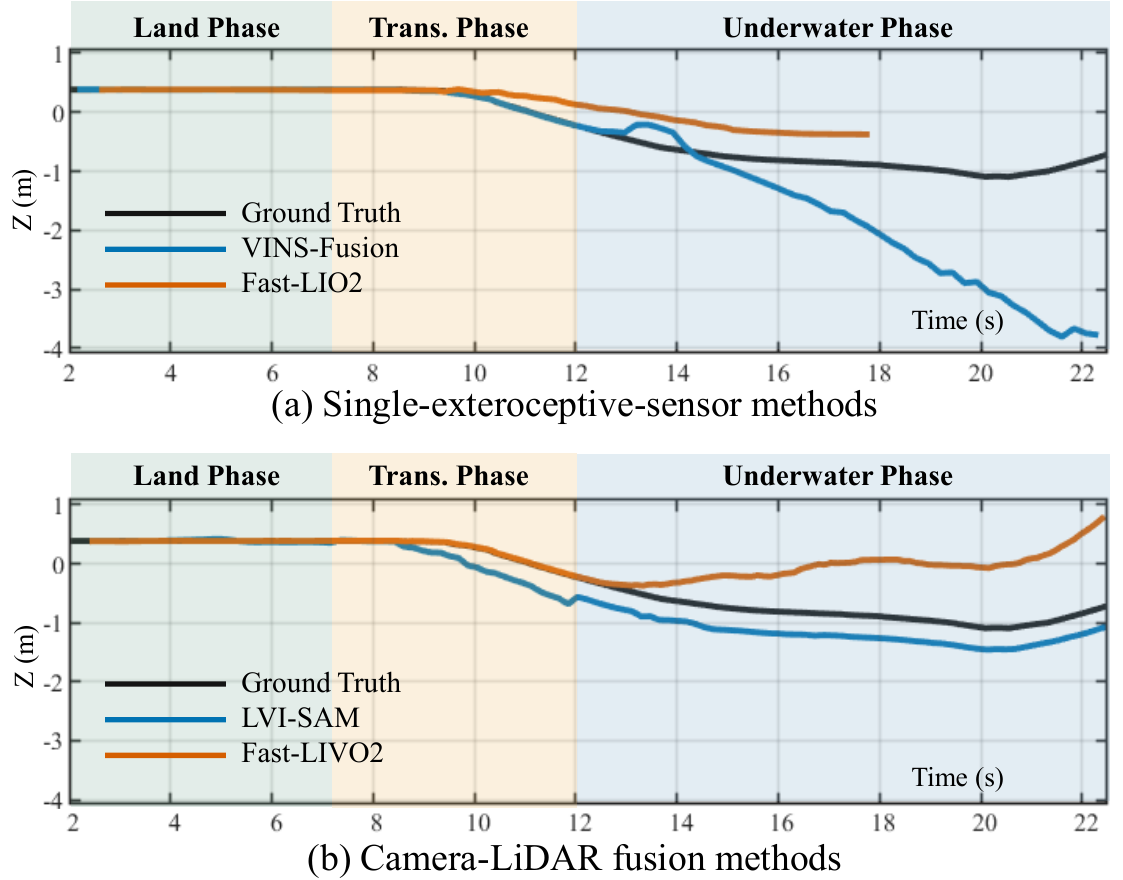}
    \caption{Comparison of vertical-position estimates during a land-to-underwater transition. (a) Methods relying on a single primary exteroceptive sensor exhibit substantial drift or tracking termination as the sensing medium changes. (b) Camera--LiDAR fusion methods maintain better trajectory continuity and reduce the localization error, but noticeable deviations remain during the transition and underwater phases. The shaded regions indicate the land, transition, and underwater phases.}
    \label{fig:case_analysis}
\end{figure}

Figure~\ref{fig:case_analysis} further illustrates how localization
errors arise and propagate during cross-domain motion. Among the
single-primary-sensor methods, Fast-LIO2 begins to deviate during the
transition phase and terminates after entering the water, indicating
that inertial propagation cannot sustain accurate estimation during
the prolonged loss of valid LiDAR constraints. VINS-Fusion produces a
complete trajectory on \texttt{seq01}, but accumulates pronounced
vertical drift during the transition and underwater phases. Continuous
pose output therefore does not necessarily imply accurate cross-domain
localization.

In comparison, the camera--LiDAR fusion methods maintain pose estimates
throughout the sequence and exhibit better overall continuity. LVI-SAM
follows the ground-truth vertical trend with a moderate offset, whereas
Fast-LIVO2 develops an increasing deviation in the later underwater
stage. These results show that multisensor fusion can alleviate
single-modality failure, but does not by itself guarantee accurate
cross-domain estimation when multiple observations degrade or become
invalid. Reliable cross-domain localization therefore requires explicit
estimation of modality reliability, rejection of invalid constraints,
and stable constraint transfer during medium transitions.

\section{Discussion}

The M-CROSS results show that abrupt changes in sensor characteristics and availability during and immediately after crossing the air--water interface remain a major challenge for continuous water--land localization. Degradation may emerge during the Transition phase or shortly after entering the target domain. Because these errors are concentrated within short intervals, full-sequence APE may dilute their effects. Continuous pose output does not necessarily imply accurate estimation, while low PW-APE may reflect limited valid coverage. TCR, PW-APE, TEE, and TSP should therefore be interpreted jointly.

Continuous water--land localization is not a simple concatenation of terrestrial and underwater SLAM. Sensors cross the interface at different times, causing asynchronous degradation and recovery. The case analysis shows that multisensor fusion can improve trajectory continuity but cannot eliminate cross-domain errors when multiple observations become invalid. Cross-domain SLAM should explicitly estimate medium states and sensor reliability, suppress unreliable constraints through dynamic weighting or gating, and smoothly transfer constraints between modalities. IMU propagation or motion models can bridge short observation gaps, while continuous state and map representations can support localization after water entry and relocalization after water exit.

M-CROSS is limited by simplified sensor models and a restricted range of platforms, scenes, trajectories, and environmental conditions. The metrics do not explicitly evaluate maximum transient error, recovery latency, or long-term consistency across repeated transitions. Therefore, M-CROSS should be regarded as a controlled, repeatable benchmark rather than a replacement for real-world water--land validation.

\section{Conclusion}

This paper presents M-CROSS, a multisensor simulation dataset and benchmark for SLAM across continuous water--land transitions. It provides LiDAR, stereo camera, IMU, imaging sonar, DVL, and depth measurements, continuous 6-DoF pose ground truth, sensor-wise medium-state annotations, and models of sensor degradation, failure, and recovery. TCR, PW-APE, TEE, and TSP evaluate trajectory completion, single-domain accuracy, transition-induced error, and robustness under prolonged interface conditions. Experiments show that multisensor fusion can improve trajectory continuity, but existing SLAM methods remain vulnerable to accuracy degradation, discontinuities, and tracking failures during medium transitions. M-CROSS provides a repeatable, unified basis for developing cross-domain state estimation and water--land SLAM methods.

\bibliographystyle{IEEEtran}
\bibliography{IEEEabrv, ref}

\end{document}